\documentclass[10pt, a4paper]{article}
\usepackage{lrec}
\usepackage{graphicx}
\usepackage{tabularx}
\usepackage{soul}
\usepackage{xcolor}
\usepackage{epstopdf}
\usepackage[T1]{fontenc}   
\usepackage[utf8]{inputenc}

\usepackage{hyperref}
\usepackage{xstring}

\usepackage[official]{eurosym}
\usepackage{titlesec}
\titlelabel{\thetitle.\quad}

\title{Language Technology Programme for Icelandic 2019--2023}

\name{\parbox{\textwidth}{\centering Anna Björk Nikulásdóttir$^1$, Jón Guðnason$^2$, Anton Karl Ingason$^3$, Hrafn Loftsson$^2$,\\Eiríkur Rögnvaldsson$^{3,4}$, Einar Freyr Sigurðsson$^4$, Steinþór Steingrímsson$^{2,4}$}}

\address{The Árni Magnússon Institute for Icelandic Studies$^4$, Grammatek$^1$, Reykjavík University$^2$, University of Iceland$^3$ \\
         Árnagarði við Suðurgötu, IS-102 Reykjavík$^{3,4}$, % ath. að á vefsíðu Árnastofnunar er vísað á skrifstofu í Árnagarði um póstfang
          Suðurgötu 57, IS-300 Akranes$^1$, Menntavegi 1, IS-102 Reykjavík$^2$ \\
         anna@grammatek.com, jg@ru.is, antoni@hi.is, hrafn@ru.is, eirikur@hi.is, \\ einar.freyr.sigurdsson@arnastofnun.is, steinthor.steingrimsson@arnastofnun.is \\}

\abstract{
In this paper, we describe a new national language technology programme for Icelandic. The programme, which spans a period of five years, aims at making Icelandic usable in communication and interactions in the digital world, by developing accessible, open- source language resources and software. The research and development work within the programme is carried out by a consortium of universities, institutions, and private companies, with a strong emphasis on cooperation between academia and industries. Five core projects will be the main content of the programme: language resources, speech recognition, speech synthesis, machine translation, and spell and grammar checking. We also describe other national language technology programmes and give an overview over the history of language technology in Iceland.  \\ \newline \Keywords{language technology programmes, infrastructure, Icelandic} }

\begin{document}

\maketitleabstract

\section{Introduction}

During the last decade, we have witnessed enormous advances in language technology (LT). Applications that allow users to interact with technology via spoken or written natural language are emerging in all areas, and access to language resources and open-source software libraries enables faster development for new domains and languages.

However, LT is highly language dependent and it takes considerable resources to develop LT for new languages. The recent LT development has focused on languages that have both a large number of speakers and huge amounts of digitized language resources, like English, German, Spanish, Japanese, etc. Other languages, that have few speakers and/or lack digitized language resources, run the risk of being left behind. 

Icelandic is an example of a language with almost a negligible number of speakers, in terms of market size, since only about 350,000 people speak Icelandic as their native language. Icelandic is therefore seldom on the list of supported languages in LT software and applications.

The Icelandic Government decided in 2017 to fund a five-year programme for Icelandic LT, based on a report written by a group of LT experts \cite{nikulasdottir_etal2017}. After more than two years of preparation, a consortium consisting of universities, institutions, associations, and private companies started the work on the programme on the $1^{st}$ of October 2019. The goal of the programme is to ensure that Icelandic can be made available in LT applications, and thus will be usable in all areas of communication. Furthermore, that access to information and other language-based communication and interaction in Icelandic will be accessible to all, e.g. via speech synthesis or speech-to-text systems.

The focus of the programme will be on the development of text and speech-based language resources, on the development of core natural language processing (NLP) tools like tokenisers, taggers and parsers, and finally, to publish open-source software in the areas of speech recognition, speech synthesis, machine translation, and spell and grammar checking. All deliverables of the programme will be published under open licenses, to encourage use of resources and software in commercial products.

While the government-funded programme for the development of resources and infrastructure software builds the backbone of the Icelandic LT programme, another branch is a competitive fund for research and development. This Strategic Research and Development Programme for Language Technology is managed by the Icelandic Centre for Research, Rannís\footnote{\url{https://rannis.is}}, which publishes calls for applications on a regular basis.

The third pillar of the programme is the revival of the joint Master's programme in LT at Reykjavik University (RU) and the University of Iceland (UI). The goal is further to increase the number of PhD students and to build strong knowledge centres for sustainable LT development in Iceland. 

The budget estimation for the programme, including the competitive fund, education plan and infrastructure costs, is around 14 million euros. Additionally, around 3.6 million euros is expected to be the contribution of the industry through the competitive fund. 

This paper is structured as follows: In Section \ref{other-programs} we discuss national LT programmes that have been run in other European countries and helped developing the Icelandic project plan. Section \ref{history} gives an overview over the 20 years of LT development in Iceland.  Section \ref{organization} shows the organisation of the new programme, and in Section \ref{projects} we describe the core projects that have been defined for it.  Finally, a conclusion is presented in Section \ref{conclusion}.

\section{Other European LT Programmes}
\label{other-programs}

In recent years, there has been much international discussion on how the future of languages depends on them being usable in the digital  world. This concern has led to a number of national LT programmes. We studied three of these national programmes: the STEVIN programme in the Netherlands which ran between 2004 and 2011, the Plan for the Advancement of Language Technology in Spain, and, in particular, the Estonian LT programmes that have been running since 2006. 

\subsection{The Netherlands}
The STEVIN programme was launched in 2004 to strengthen the position of Dutch in LT by building essential resources for the language. Its objectives were to raise awareness of LT in order to stimulate demand for LT products, to promote strategic research in the field and develop essential resources, and to organise the management, maintenance and distribution of language resources that have been developed \cite{dhalleweyn-etal-2006-dutch}. The programme was based on cooperation between government, academia and industry, both in Flanders and the Netherlands. It encompassed a range of projects from basic resources to applications for language users, and attention was paid to distribution, dissemination and valorisation of project results by means of the HLT Agency, which also had a role in clearing intellectual property rights (IPRs) and issuing licence agreements \cite{Boekestein2006FunctioningOT}. 
 
%The programme was a coordinated effort by a number of Dutch and Flemish institutions. The funding parties, together with experts in the field, formed a programme board. The board supervised the programme and made decision on grants and a special programme committee,  which planned the programme, ensured that its policies were enforced. The committee wrote calls for proposals and advised the board on funding decisions, after the proposals were assessed independently by an international advisory panel. A special LT programme office, which was run by two institutions, was responsible for project management \cite{dhalleweyn-etal-2006-dutch}.

The general targets of the STEVIN programme were reached to a large extent. According to a report on the results of the programme \cite{Spyns2013}, it resulted in a network with strong ties between  academia  and  industry, beneficial  for  future  utilisation  of the  STEVIN  results. The evaluators of the programme qualified it as successful, but had recommendations for a future programme, if initiated. They suggested more interaction with other similar (inter)national R\&D programmes, asserted that the complexity of IPR issues had been seriously underestimated and called for a better clarification of the role of open-source. The total cost of the STEVIN programme was over 10 million euros, of which well over 80\% was spent on R\&D projects.

\subsection{Spain}
The Spanish LT programme Plan for Advancement of Language Technology started in 2016, and is scheduled to finish in 2020. Its aims are to develop infrastructure for LT in Spain, specifically for Spanish and the co-official languages, Basque, Catalan, Galician and Aranese. Furthermore, to promote the LT industry by boosting knowledge transfer between research and industry actors, and to improve the quality and capacity of public services by employing NLP and machine translation (MT) technology. Government should be the leading participant in LT with high-profile projects in  healthcare, as well as in the judicial and educational systems, and in tourism \cite{spanishplan}.

%The programme was prepared by a steering committee with the participation of a number of state secretariats and other governing bodies, which in turn created a committee of experts from academia, key companies in the Spanish ICT sector and those specialised in NLP and MT, as well as representatives of government entities.

The plan was to facilitate the development of tools and linguistic resources. Examples of tools are named entity recognisers, word-sense disambiguation, tools for computing semantic similarity and text classification, automatic summarisation and MT. Examples of linguistic resources to be developed in the programme are parallel corpora, lists of proper nouns, terminology lists and dictionaries. 

The estimated total cost of the programme was 90 million euros. As the programme had just recently started when the Icelandic programme was being planned, we did not have any information on what went well and what could have been done better.

\subsection{Estonia}
Regarding LT, the Estonian situation is, in many ways, similar to that of Iceland: It has too few users for companies to see opportunities in embarking on development of (costly) LT, but on the other hand society is technologically advanced -- people use, or want to be able to use, LT software. In Estonia, the general public wants Estonian to maintain its status, and like Icelandic, the language has a complex inflection system and very active word generation. The problems faced by Estonia are therefore not unlike those that Iceland faces.

In Estonia, three consecutive national programmes have been launched. The third national programme, \emph{Estonian Language Technology 2018--2027}, is currently under way. While the Estonian Ministry of Education and Research has been responsible for the programmes, the universities in Tallinn and Tartu, together with the Institute of the Estonian Language, led the implementation. 

%When planning the Icelandic LT programme, the third programme had not yet started so we considered the results of the first one and the planning and operation of the second one.

The National Programme for Estonian Language Technology was launched in 2006. The first phase ran from 2006 to 2010. All results of this first phase, language resources and software prototypes, were released as public domain. All such resources and tools are preserved long term and available from the Center of Estonian Language Resources. 33 projects were funded, which included the creation of reusable language resources and development of essential linguistic software, as well as bringing the relevant infrastructure up to date \cite{vider12}. The programme managed to significantly improve upon existing Estonian language resources, both in size, annotation and standardisation. In creating software, most noticeable results were in speech technology. %Considerable improvements were made in speech synthesis and a speech recognizer was developed for Estonian. 
Reporting on the results of the programme \cite{vider12} stress that the first phase of the programme created favourable conditions for LT development in Estonia. According to an evaluation of the success of the programme, at least 84\% of the projects had satisfactory results. The total budged for this first phase was 3.4 million euros.

The second phase of the programme ran from 2011 to 2017 with a total budget of approx. 5.5 million euros. It focused on the implementation and integration of existing resources and software prototypes in public services. Project proposals were called for, funding several types of actions in an open competition. %: R\&D to create new software, new language resources, Center for Estonian Language Resources, Integration of LT into other applications, and specially aimed projects on the order of the steering committee. 
The main drawback of this method is that it does not fully cover the objectives, and LT support for Estonian is thus not systematically developed. Researchers were also often mostly interested in results using prototypes rather than stable applications. As most of the projects were instigated at public institutes, relation to IT business was weak. Furthermore, the programme does not deal explicitly with LT education. On the other hand, the state of LT in Estonia soon become relatively good compared to languages with similar number of speakers \cite{vider15}.

\section{History of Icelandic LT}
\label{history}
The history of Icelandic LT is usually considered to have begun around the turn of the century, even though a couple of LT resources and products were developed in the years leading up to that. Following the report of an expert group appointed by the Minister of Education, Science and Culture \cite{tungutaekni_1999}, the Icelandic Government launched a special LT Programme in the year 2000, with the aim of supporting institutions and companies to create basic resources for Icelandic LT work. This initiative resulted in a few projects which laid the ground for future work in the field. The most important of these were a 25 million token, balanced, tagged corpus, a full-form database of Icelandic inflections, a training model for PoS taggers, an improved speech synthesiser, and an isolated word speech recogniser \cite{rognvaldsson_2008}.

After the LT Programme ended in 2004, researchers from three institutions, UI, RU, and the Árni Magnússon Institute for Icelandic Studies (AMI), joined forces in a consortium called the \textit{Icelandic Centre for Language Technology (ICLT)}, in order to follow up on the tasks of the Programme. In the following years, these researchers developed a few more tools and resources with support from The Icelandic Research Fund, notably a rule-based tagger, a shallow parser, a lemmatiser, and a historical treebank \cite{helgadottir_2013}. 

In 2011--2012, researchers from the ICLT also participated in two speech technology projects initiated by others:  A new speech synthesiser for Icelandic which was developed by the Polish company Ivona, now a subsidiary of Amazon, for the Icelandic Association for the Visually Impaired, and a speech recogniser for Icelandic developed by Google \cite{helgadottir_2013}. 

Iceland was an active participant in the META-NORD project, a subproject of META-NET\footnote{\url{http://meta-net.eu}}, from 2011 to 2013. Within that project, a number of language resources for Icelandic were collected, enhanced, and made available, both through META-SHARE and through a local website, \url{málföng.is} (\textit{málföng} being a neologism for `language resources'). Among the main deliveries of META-NET were the Language White Papers \cite{rehm_2012}. The paper on Icelandic \cite{rognvaldsson_2012} highlighted the alarming status of Icelandic LT. Icelandic was among four languages that received the lowest score, ``support is weak or non-existent'' in all four areas that were evaluated. 

The White Paper received considerable attention in Icelandic media and its results were discussed in the Icelandic Parliament. In 2014, the Parliament unanimously accepted a resolution where the Minister of Education, Science and Culture was given mandate to appoint an expert group which should come up with a long-term LT plan for Icelandic. The group delivered its report to the Minister in December 2014. The result was that a small LT Fund was established in 2015.

During the last years, a strong centre for speech technology has been established at RU, where development in speech recognition and synthesis has been ongoing since 2011. Acoustic data for speech recognition was collected and curated at RU~\cite{gudhnason2012almannaromur,petursson2016eyra,steingrimsson2017malromur} and a baseline speech recognition system for Icelandic was developed~\cite{nikulasdottir2018open}.  Specialised speech recognisers have also been developed at RU for the National University Hospital and Althingi~\cite{helgadottir2017building,helgadottir2019althingi,runarsdottir2019lattice}.  A work on a baseline speech synthesis system for Icelandic has also been carried out at RU~\cite{nikulasdottir2018icelandic,nikulasdottir2019bootstrapping}.

The AMI has built a 1.3-billion-word corpus, the Icelandic Gigaword Corpus (IGC) \cite{steingrimsson_2018}, partially funded by the Icelandic Infrastructure Fund. Further, a private company, Miðeind Ltd., has developed a context-free parser \cite{thorsteinsson_2019} partially funded by the LT Fund.

In October 2016, the Minister of Education, Science and Culture appointed a special LT steering group, consisting of representatives from the Ministry, from academia, and from the Confederation of Icelandic Enterprise (CIE). The steering group commissioned three LT experts to work out a detailed five-year Project Plan for Icelandic LT. The experts delivered their proposals, \textit{Language Technology for Icelandic 2018--2022 -- Project Plan} \cite{nikulasdottir_etal2017} to the Minister in June 2017.

\section{Organisation of the Icelandic LT Programme 2019--2023}
\label{organization}
The Icelandic Government decided soon after the publication of the report \textit{Language Technology for Icelandic 2018--2022 -- Project Plan} to use the report as a base for a five-year government funded LT programme for Icelandic. The self-owned foundation \textit{Almannarómur}, founded in 2014 to support the development of Icelandic LT, was to be prepared to take over a role as a Centre of Icelandic LT and to elaborate on how the programme could be organised and executed to meet the goals defined in the report.

The Icelandic Ministry of Education, Science and Culture signed an agreement with Almannarómur in August 2018, giving Almannarómur officially the function of organising the execution of the LT programme for Icelandic. Following a European Tender published in March 2019, Almannarómur decided to make an agreement with a consortium of universities, institutions, associations, and private companies (nine in total) in Iceland (listed in Table \ref{sim}) to perform the research and development part of the programme. This Consortium for Icelandic LT (\textit{Samstarf um íslenska máltækni -- SÍM}) is a joint effort of LT experts in Iceland from academia and industry. SÍM is not a legal entity but builds the cooperation on a consortium agreement signed by all members.
During the preparation of the project, an expert panel of three experienced researchers from Denmark, the Netherlands, and Estonia was established to oversee the project planning and to evaluate deliverables at predefined milestones during the project.

\begin{table}[!h]
\begin{center}
\begin{tabularx}{\columnwidth}{|l|X|}

      \hline
      SÍM member&Website\\
      \hline
      The Árni Magnússon Instit. & \\ for Icelandic Studies & https://arnastofnun.is\\
      \hline
      Reykjavik University (RU) & https://www.ru.is\\
      \hline
     University of Iceland (UI) & https://www.hi.is\\
     \hline 
     RÚV & https://www.ruv.is\\
      \hline
      Creditinfo & https://www.creditinfo.is\\
      \hline
      The Association of the & \\Visually Impaired & https://www.blind.is\\
      \hline
      Grammatek & https://grammatek.com\\
      \hline
      Miðeind & https://mideind.is\\
      \hline
      Tiro & https://tiro.is\\
      \hline

\end{tabularx}
\caption{Members of the SÍM consortium for Icelandic LT}
\label{sim}
 \end{center}
\end{table}

SÍM has created teams across the member organisations, each taking charge of a core project and/or defined subtasks. This way the best use of resources is ensured, since the team building is not restricted to one organisation per project. One project manager coordinates the work and handles communication and reporting to Almannarómur and the expert panel.

Besides the role of the executive of the research and development programme itself, Almannarómur will conduct communication between the executing parties and the local industry, as well as foreign companies and institutions. Together with the executing parties, Almannarómur will also host conferences and events to promote the programme and bring together interested parties.

\section{Core Projects}
\label{projects}
In this section, we describe the five core projects that have been defined in the Icelandic LT programme.

\subsection{Language Resources}
\label{section:lr}
As mentioned above, a number of language resources have been made available at the repository \textit{málföng}.\footnote{\url{http://málföng.is/}} Most of these are now also available at the CLARIN-IS website\footnote{\url{https://clarin.is/en/resources/}} and will be integrated into the CLARIN Virtual Language Observatory.\footnote{\url{https://www.vlo.clarin.eu/}} Below we give a brief and non-exhaustive overview of language resources for Icelandic which will be developed in the programme.

%\subsubsection{Tagged Corpora}

%\begin{itemize}
\begin{enumerate}
    \item \textbf{Tagged corpora}.
The IGC \cite{steingrimsson_2018} contains 1.3 billion running words, tagged and lemmatised. It is much bigger than previous tagged corpora, most notably the Icelandic Frequency Dictionary (IFD; \nocite{pind1991}Pind et al., 1991), which was the first morphologically tagged corpus of Icelandic texts, containing half a million words tokens from various texts, and the Tagged Icelandic Corpus (MÍM; \nocite{helgadottir_2012mim}Helgadóttir et al,. 2012), a balanced corpus of texts from the first decade of the 21st century, containing around 25 million tokens. A gold standard tagged corpus was created from a subset of MÍM \cite{Loftsson2010DevelopingAP}. Some revisions of the morphosyntactic tagset used for tagging Icelandic texts will be done in the programme, and the gold standard updated accordingly.

We will update the IGC with new data from more sources and continue collecting data from rights holders who have given their permission for using their material. A new version will be released each year during the five-year programme.
%\end{itemize}

%The Icelandic Frequency Dictionary (IFD; \nocite{pind1991}Pind et al., 1991) was the first morphologically tagged corpus of Icelandic texts. It contains a half a million words from various texts, used for the building of a frequency dictionary. The tokens are both lemmatized and tagged for parts of speech and morphological features. The Tagged Icelandic Corpus (MÍM; Helgadóttir et al.) is, however, only morphologically tagged but not lemmatized. It is centered around contemporary texts (from the first decade of the 21st century) and is a lot bigger than the IFD, containing around 25 million tokens. The largest Icelandic corpus, on the other hand, is the Icelandic Gigaword Corpus (IGC; \nocite{steingrimsson_2018}Steingrímsson et al., 2018), containing 1.3 billion running words, which are both tagged and lemmatized.  

%- safna gögnum frá fleirum
%- halda áfram að safna frá þeim sem hafa gefið leyfi
%This corpus is intended to be constantly updated. Once the first version has been released, data collection needs to continue from the rights holders who have given their permission.

%\subsubsection{Treebanks}

%\begin{itemize}
    \item \textbf{Treebanks}.
The largest of the syntactically parsed treebanks that exist is the Icelandic Parsed Historical Corpus (IcePaHC; \nocite{rognvaldsson-etal-2012-icelandic,icepahc09,rognvaldsson2011creating}Wallenberg et al., 2011; R\"o{}gnvaldsson et al., 2011, 2012), which contains one million words from the 12$^{th}$ to the 21$^{st}$ century. The scheme used for the syntactic annotation is based on the Penn Parsed Corpora of Historical English \cite{kroch00,kroch04}. On the other hand, no Universal Dependencies (UD)-treebanks are available for Icelandic. Within the programme, a UD-treebank will by built, based on IcePaHC, and extended with new material.
%\end{itemize}

%\subsubsection{A Morphological Database}

%\begin{itemize}
    \item \textbf{Morphological database}.
The Database of Icelandic Morphology (DIM; \nocite{bjarnadottir-etal-2019-dim}Bjarnadóttir et al., 2019) contains inflectional paradigms of about 287,000 lemmas. %%%
A part of the database, DMII-Core, only includes data in a prescriptive context and is suited for language learners, creating teaching material and other prescriptive uses. It consists of the inflection of approx.~50,000 words. % in the Icelandic Gigaword Corpus.  
We will extend it by reviewing ambiguous inflection forms. We will define format for data publication as the core will be available for use by a third party. For the sake of simplifying the process of adding material to the database and its maintenance, we will take advantage of the lexicon acquisition tool described in Section \ref{nlptools} and adapt it for DIM.
%which takes into account typographical errors and inflectional variation. Such data can be used in LT research even though it will not necessarily be published in the database.
%and is the linchpin of the Database of Icelandic Morphology (DIM) which also contains the DMII Core and MorphIce. The core consists of the inflection of the 50,000 most frequent words in the Icelandic Gigaword Corpus and can be used for a third party publication through an API. 
%MorphIce, on the other hand, gives a morphological analysis (binary constituent structure) of the words (lemmas) and lemmatization of each constituent.

%- extend the core by reviewing ambiguous inflection forms - the core will be available for use by a third party... 
%- define format for data publication
%- adapt a lexicon acquisition tool
%[- MorphIce - unnið upp úr þessu
%- data model allows for prentvillur, stafsetningarvillur og tilbrigði, án þess að þau séu birt á vefnum en hægt að nota í LT, sem ekki var hægt áður]

\item \textbf{Hyphenation tool}. Hyphenation from one language to another often seems rather idiosyncratic but within one and the same language, such as Icelandic, such rules are often reasonably clear. A list of more than 200,000 Icelandic words with permissible hyphenations is available in the language resources repository. It will be expanded based on words from the DIM. A new hyphenation tool, trained on the extended list, will be built in the programme. The tool makes a suggestion for correct hyphenation possibilities of words that are not found on the hyphenation list. %list will be the basis for a tool that hyphenates words and will be used to train a tool for words not on the list, the new tool will guess the correct hyphenation possibilities.

\item \textbf{Icelandic wordnet}. 
The Icelandic wordnet \cite{ordanet2011}, which contains 200,000 phrasemes of various kinds and about 100,000 compounds, is not a traditional dictionary as it analyses internal connections semantically and syntactically within Icelandic vocabulary. We will define a more appropriate data format and convert the wordnet data to that format. In addition, we will work on improving the wordnet itself by filling in gaps in various categories. %However, the wordnet data needs to be converted to a more appropriate data format, which we will define. 

%íslenskt orðanet ... Icelandic Wordnet (\nocite{ordanet2011}Jónsson and Úlfarsdóttir, 2011), which contains 200,000 phrasemes of various kinds and about 100,000 compounds, is monolingual. It is not a traditional dictionary as it analyses internal connections semantically and syntactically within Icelandic vocabulary.
%\end{itemize}
\end{enumerate}

\subsection{NLP Tools}
\label{nlptools}
A wide variety of NLP tools are to be developed or improved upon within the programme. It is of vital importance to develop quality NLP tools, as many tools often form a pipeline that analyses data and delivers the results to tools used by end users, and, in the pipeline, errors can accumulate and perpetuate. 

When the programme started, there were a few available tools for Icelandic. IceNLP \cite{Loftsson2007IceNLPAN} is a suite of NLP tools containing modules for tokenisation, PoS-tagging, lemmatising, parsing and named entity recognition. Greynir \cite{thorsteinsson_2019} is a full parser which also includes a tokeniser and recognises some types of named entities. Nefnir \cite{ing19} is a lemmatiser which uses suffix substitution rules, derived from the Database of Icelandic Morphology \cite{bjarnadottir-etal-2019-dim}, giving results that outperform IceNLP. ABLTagger \cite{ste19} is a PoS tagger that outperforms other taggers that have been trained for tagging Icelandic texts.

Some of these tools give good results, but can be improved upon. For other tasks, new tools need to be built. As part of the release process care will be taken to ensure all resulting software are up to high quality standards, and well documented to facilitate use by third parties. Where applicable, RESTful APIs will also be set up to further promote the usage of the products.

%\textcolor{red}{
%ABN: as one might get the impression when reading the NLP section, that there is not much to be done since most of the tools have already been developed for Icelandic, maybe add something on the software side, i.e. that the publishing and packaging all software as robust and usable APIs / modules is also an important part of the work to evolve the tools from the prototype status?}

\begin{enumerate}
\item \textbf{Tokeniser}.
%\subsubsection{Tokeniser}
A basic step in NLP is to segment text into units, normally sentences and tokens. Since any errors made at this stage will cascade through the process, it is important that the tokeniser is as accurate as possible. %The main challenge faced in sentence segmentation is to decide whether a full-stop means the end of a sentence and whether a capital letter denotes the start of a sentence. In this respect,
%Icelandic faces challenges similar to those of other languages, but  
A tokeniser for Icelandic needs to be able to correctly recognises abbreviations, time units, dates, etc.  It must also be adjustable and able to run using different settings, since its  output must be adaptable to different projects and different uses.

Previously, two tokenisers have been built for Icelandic, one is a part of IceNLP and the other a part of Greynir. As Greynir is still in active development, it will be used as a base for the LT project's development.
%it is probably the best option to develop the Greynir tokeniser further.
In order to be able to test the tokenisers' accuracy, a test set that takes different tokeniser settings into account will be developed.

\item \textbf{PoS tagger}.
%\subsubsection{PoS Tagger}
Precise PoS-tagging is important in many LT projects because information on word class or morphological features is often needed in later stages of an NLP pipeline. Improved tagging accuracy, thus often results in an improvement in the overall quality of LT software. 

A number of PoS-taggers have been developed for Icelandic, with the best results achieved by a recent bidirectional LSTM tagging model \cite{ste19}.
%It is important to keep improving  tagging  methods  for  Icelandic,  and  to  analyse  how  far it is possible to progress, both theoretically (how similar the results would be if two capable linguists were to tag the same text) and technically. [Myndi sleppa þessu: HL]
While developing PoS taggers for Icelandic further using state-of-the-art methods, we will also study and try to estimate how much accuracy can theoretically be reached in tagging a variety of Icelandic text styles, using the tag set chosen for the LT programme (see Section \ref{section:lr}).

\item \textbf{Lemmatiser}.
%\subsubsection{Lemmatiser}
A new lemmatiser for Icelandic, Nefnir, has recently been published \cite{ing19}. It has been shown to be quite accurate, although a standardised test set is not available to compare it to other lemmatisers, like Lemmald \cite{ingason2008mixed}. Its main weakness is in lemmatising unknown words, which is a hard problem for inflected languages. We will study if its accuracy can be improved in that regard.

\item \textbf{Parser}.
%\subsubsection{Parser}
Three parsers have previously been developed for Icelandic. IceNLP includes a shallow parser based on a cascade of finite-state transducers \cite{Loftsson2007IceParser}. Greynir, on the other hand, fully parses sentences according to a hand-crafted context-free grammar. 
%It generates all possible parse trees for a sentence and select a single best one as output based on a special scoring heuristic. 
A parsing pipeline for Icelandic based on the IcePaHC corpus and the Berkeley-parser has also been released \cite{jokulsdottir2019parsing}. %{The pipeline is available on \url{https://github.com/antonkarl/iceParsingPipeline.}} 
No Universal Dependencies (UD) parser is available for Icelandic and no UD treebank, but in a project that started in 2019, independent of the LT programme, IcePaHC \cite{rognvaldsson-etal-2012-icelandic} will be converted to a UD treebank. 

The IceNLP and Greynir parsers will be evaluated and either one of them or both developed further. We will also adapt a UD-parser to Icelandic UD-grammar.

\item \textbf{Named entity recogniser}.
%\subsubsection{Named Entity Recogniser}
Some  work  has  been  carried  out  on  named  entity  recognition  for  Icelandic. IceNLP contains a rule-based module that has achieved 71-79\% accuracy and a recent tool based on a bidirectional LSTM \cite{ingolfsdottir-etal-2019-towards} obtained an F1 score of 81.3\%. There is also a named entity recogniser for proper names in Greynir, but its accuracy has not yet been evaluated. Within the programme, different training methods will be experimented with and evaluated, and the most promising tools evaluated further.

\item \textbf{Semantic analysis}.
%\subsubsection{Semantic Analysis}
A variety of different tasks involve semantic analysis, including word-sense disambiguation (WSD), anaphora  resolution, identifying co-references, analysing semantic similarity between compound verbs and phrases, and more.
%assessing the semantic relations between words or similarity in the meanings of  words or phrases, and detecting semantic roles. It is also possible to use semantic analysis to extract semantic information from a large number of texts to create or add to semantic databases, such as wordnets. 

We will work on these four aspects of semantic analysis listed above. In recent years, not much work has been carried out in this field for Icelandic. This part of the LT programme will thus start with researching the current state-of-the-art and defining realistic goals. 

%\subsubsection{Annotation tools}
%Texts often need manual annotation to some, or all, extent when they are being pre-processed for training and/or developing LT tools. Texts can be annotated with grammatical and semantic information; individual aspects, such as person names,  place  names  or  dates;  writing  errors,  etc.  Annotation,  manual  or  automatic, is carried out to enable other software to read information from standardised labels.

%It is important that annotations can be carried out in such a way that annotated data is standardized and can be input into schemes to annotate it further. This facilitates use and reuse of data in different systems, which is important to get as much as possible out of labour intensive manual work. 

%We will define requirements for different annotation projects and choose open-source software where available. [ABN: kannski óþarfi að fjalla um ]

\item \textbf{Lexicon acquisition tool}.
%\subsubsection{Lexicon Acquisition Tool}
When constructing and maintaining lexical databases, such as DIM, the Icelandic  wordnet or other related resources, it is vital to be able to systematically add neologies and words that are missing from the datasets, especially those commonly used in the language.  %The editors of the lexical resources can use such a tool to find gaps in their data, monitor when new words gain a foothold in the language, and see in what context the words are used. Besides being a useful tool for helping with building necessary resources, a lexicon acquisition tool can also be useful in composing dictionaries and in research on modern language. Such a tool needs to be flexible and adjustable to the needs of individual resources. 
Within the LT programme a flexible lexicon acquisition tool will be developed. It will be able to identify and collect unknown words and word forms, together with statistics, through structured lexical acquisition from the Icelandic Gigaword Corpus, which is constantly being updated, and other data sources in the same format.
\end{enumerate}

\subsection{Automatic Speech Recognition (ASR)}
The main aim of the automatic speech recognition (ASR) project is to gather all necessary language and software resources to implement and build standard speech recognition systems for Icelandic.  The project should enable developers to either research, develop or implement ASR without having to gather language resources.  To achieve this goal, the project is divided into data gathering, recipe development, and software implementation and research.

\begin{enumerate} 
\item \textbf{Data gathering}.
The data gathering part of the project encompasses a wide variety of speech and transcript resources. A continuation of the Málrómur project~\cite{steingrimsson2017malromur} has already  been implemented using Mozilla Common Voice\footnote{The project is called \emph{Samrómur} and is accessible here: \url{https://samromur.is/}}. Here the aim is to double the size of the existing data set, get a more even distribution of speakers across geographic locations and age groups, and gather data from second language speakers. Additionally, radio and television transcripts are being gathered on a large scale and prepared for publication for ASR development.  Conversations, queries and lectures will also be transcribed and published, and large open historical data sets will be aligned and prepared for publication.

\item \textbf{Recipe development}.
ASR recipes for Icelandic will be developed further using more language resources~\cite{nikulasdottir2018open} and specific application areas such as conversations, question answering and voice commands will be given a special attention.  ASR systems that focus on teenagers, children and second language speakers are also within the scope of the project.  These recipes are then used to create resources for smart-phone and web-based integration of ASR for Icelandic.

\item \textbf{Software implementation and research}.
The research areas are chosen so to enhance the language resource development for Icelandic. A punctuation system for Icelandic will be analysed and implemented.  Compound words are common in Icelandic and the language also has a relatively rich inflection structure so it is important to address those features for language modeling.  Pronunciation analysis, speaker diarization and speech analysis will also be addressed especially for Icelandic, and acoustic modelling for children and teenagers receive attention in the project.
\end{enumerate}

\subsection{Speech Synthesis (TTS)}. 
The text-to-speech project will produce language resources that enable voice building for Icelandic. 
\begin{enumerate}
\item \textbf{Unit selection}.
Eight voices for unit-selection TTS will be recorded, with the aim of attaining diversity in age and dialect, with an equal number of male and female voices.
%Eight unit-selection and forty statistical parametric speech synthesis (SPSS) voices will be recorded during the project.
The reason why unit-selection is chosen is to increase the likelihood that the project will produce useful and viable voices that can be used in addition to the two unit-selection voices that already exist for Icelandic. 
\item \textbf{Statistical parametric speech synthesis}. 
Forty voices for statistical parametric speech synthesis (SPSS) will be recorded during the project. The plan is to publish open-source unit-selection and SPSS recipes with all necessary language resources so that programmers and researchers can continue to develop voices for Icelandic.

Suitable TTS voices for web-reading and smartphones will be developed within an open-source paradigm.  This will allow the industry to use the voices developed within the project.  
\item \textbf{Research}. 
The targeted research part of the project will facilitate the recipe development and software implementation.  Quality assessment systems will be set up, text normalization for Icelandic will be developed fully, and intonation analysis for Icelandic will be implemented and applied to TTS.
\end{enumerate}

\subsection{Spell and Grammar Checking}

The Spell and Grammar Checking project will develop and make freely available, under open-source licensing, important data sets and tools for further establishment of automated text correction systems for Icelandic. The project makes extensive use of other resources that have been developed independently, or will be developed within the larger framework of the current LT Programme for Icelandic, including the Database of Icelandic Morphology \cite{bjarnadottir-etal-2019-dim}, the Greynir system \cite{thorsteinsson_2019}, and the Icelandic Gigaword corpus \cite{steingrimsson_2018}. On the one hand, the project focuses on developing error corpora for Icelandic, and on the other, it focuses on creating a set of correction tools. Challenges associated with richly inflected languages continue to be a matter of central interest in this project, like previous work on Icelandic spelling correction \cite{ingason2009context}.
% AKI: hér væri gott að vitna líka í Skramba en veit ekki um neina grein um hann

\begin{enumerate}
    \item \textbf{Text correction data}. The data construction aspect of the project will develop three error corpora that can be used for quantitative analysis of errors in written Icelandic text. The error corpora will also serve as a foundation for training data-driven training correction systems. One corpus will focus on the written language of Icelandic speakers who are not known to have unusual language properties. Another corpus will focus on speakers who are in the process of learning Icelandic as a second language, and a third one will include data from dyslexic speakers.
    
    \item \textbf{Software development}. The software development tasks of the spell and grammar checking project will build a working open source correction system whose development is informed by the analysis of the data sets created within the project. The spell and grammar checker will be based on the foundation for processing Icelandic text provided by the Greynir system.
\end{enumerate}

\subsection{Machine Translation}
The purpose of the MT project is to build open-source systems capable of translating between Icelandic and English, in both directions, is$\rightarrow$en and en$\rightarrow$is. The goal is that the translation quality will be good enough to be useful for translators in specific domains. A part of the MT project is indeed to define in which translation domain most value can be gained with the systems.

Very limited work on MT for Icelandic has been carried out since the turn of the century. A prototype of an open-source is$\rightarrow$en rule-based MT system has been developed using the Apertium platform \cite{brandt_2011}, but this system is not currently in public use.

The AMI has recently compiled an English-Icelandic parallel corpus, \textit{ParIce}, the first
parallel corpus built for the purposes of
MT research and development for Icelandic \cite{barkarson_2019}. The primary goal of the compilation of ParIce was to build a corpus large enough and of good enough quality for training useful MT systems.  ParIce currently consists of 39 million Icelandic words
in 3.5 million segment pairs. The largest parts of ParIce consists of film and TV subtitles from the Opus corpus \cite{tiedemann_2012}, and texts from the European Medicines Agency document portal, included in the Tilde MODEL corpus \cite{rozis_2017}.

%The company Miðeind Ltd. has experimented with developing is$\rightarrow$en and en$\rightarrow$is systems based on neural networks trained on Parice, but theses systems are also not currently in public use.

Google Translate\footnote{\url{https://translate.google.com/}} supports translations between Icelandic and various languages and is currently used widely by Icelanders and foreigners for obtaining understandable translations of  given texts (the task of assimilation). The problem with Google's system is, however, that neither the source code nor the training data is publicly available. Moreover, the system is a general translation engine, but not developed specifically for translating texts in a particular domain.

Our MT project in the new LT programme consists of the following sub-parts:
\begin{enumerate}
    \item \textbf{Parallel data}. Icelandic's rich morphology and relatively free word order is likely to demand large amount of training data in order to develop MT systems that produce adequate and fluent translations. The ParIce corpus currently consists of only 3.5 million sentence pairs which is rather small in relation to parallel corpora in general. The goal of this phase is to create an aligned and filtered parallel corpus of translated documents from the European Economic Area (EEA) domain (e.g. regulations and directives). As of 2017, around 7,000 documents were available in Icelandic with corresponding documents in English. The aim is to pair all accessible documents in the course of the project.
    \item \textbf{Back-translation}. In order to augment the training data, back-translated texts will be used. Monolingual Icelandic texts will be selected and translated to English with one of the baseline system (see below). By doing so, more training data can be obtained for the en$\rightarrow$is direction. An important part of using back-translated texts during training is filtering out translations that may otherwise lead to poor quality of the augmented part.
    \item \textbf{Baseline system}. In this part, three baseline MT systems will be developed. First, a statistical phrase-based MT system based on Moses \cite{koehn-2007}, second, a bidirectional LSTM model using the neural translation system OpenNMT \cite{klein-2017}, and, third, a system based on an attention-based neural network \cite{bahdanau_2015} using Tensor2Tensor\footnote{\url{https://github.com/tensorflow/tensor2tensor}}. All the three systems will be trained on ParIce, and the additional data from tasks 1 and 2 above. Eventually, the goal is to choose the best performing MT-system for further development of MT for Icelandic.
    \item \textbf{MT interface}.
    An API and a web user interface for the three baseline systems, mentioned in item 3 above, will be developed to give interested parties access to the systems under development, and to establish a testing environment in which members of the public can submit their own text. Thus, results from the three systems can be compared directly, as well as to the translations produced by Google Translate. Moreover, in this part, a crowd-sourcing mechanism will be developed, i.e. a functionality to allow users to submit improved translations back to the system for inclusion in the training corpus.
\item \textbf{Pre- and postprocessing}. Preprocessing in MT is the task of changing the training corpus/source text in some manner for the purpose of making the translation task easier or mark particular words/phrases that should not be translated. Postprocessing is then the task of restoring the generated target language to its normal form. An example of pre- and postprocessing in our project is the handling of named entities (NEs). NEs are found and matched within source and target sentence pairs in the training corpus, and replaced by placeholders with information about case and singular/plural number. NE-to-placeholder substitution is implemented in the input and placeholder-to-NE substitution in the output pipelines of the translation system.
\end{enumerate}

\section{Conclusion}
\label{conclusion}

%We have described a new, five-year, LT programme for Icelandic where the focus is on developing various language resources and tools, and publishing open-source software. We can learn a lot from somewhat similar national programmes in other countries. Despite the steep path to getting where we are, we have every reason to be optimistic about the future of Icelandic LT.

We have described a five-year, national LT programme for Icelandic. The goal is to make Icelandic useable in communication and interactions in the digital world. Further, to establish graduate and post-graduate education in LT in Iceland to enable the building of strong knowledge centres in LT in the country.

After studying somewhat similar national programmes in other European countries, we have defined the most important factors that in our opinion will help lead to the success of the programme: First, we have defined core projects that comprise the most important language resources and software tools necessary for various LT applications. Second, all deliverables will be published under as open licenses as possible and all resources and software will be easily accessible. The deliverables will be packaged and published for use in commercial applications, where applicable. Third, from the beginning of the programme, we encourage innovation projects from academia and industry through a competitive R\&D fund, and fourth, constant communication with users and industry through conferences, events and direct interaction will be maintained, with the aim of putting deliverables to use in products as soon as possible. The cooperation between academia and industry is also reflected in the consortium of universities, institutions, associations, and private companies, performing the R\&D work for all core projects. 

The described plan is tied in with 20 years of LT history in Iceland, and despite the steep path to getting where we are, we have every reason to be optimistic about the future of Icelandic LT.

\section{Bibliographical References}
\label{main:ref}

\bibliographystyle{lrec}
\bibliography{LTProgramIcelandic}

\end{document}